\documentclass[nohyperref]{article}

\usepackage{microtype}
\usepackage{graphicx}
\usepackage{subfigure}
\usepackage{booktabs} 

\usepackage{hyperref}

\usepackage[accepted]{icml2022}

\usepackage{amsmath}
\usepackage{amssymb}
\usepackage{mathtools}
\usepackage{amsthm}

\usepackage[capitalize,noabbrev]{cleveref}

\theoremstyle{plain}

\theoremstyle{definition}

\theoremstyle{remark}

\usepackage[textsize=tiny]{todonotes}

\icmltitlerunning{OOD Detection in Graph Classification}

\usepackage[justification=centering]{caption}
\begin{document}

\twocolumn[
\icmltitle{Towards OOD Detection in Graph Classification \\ from Uncertainty Estimation Perspective}



\icmlsetsymbol{equal}{*}

\begin{icmlauthorlist}
\icmlauthor{Gleb Bazhenov}{Skoltech}
\icmlauthor{Sergei Ivanov}{Skoltech,Criteo}
\icmlauthor{Maxim Panov}{TII,Skoltech}
\icmlauthor{Alexey Zaytsev}{Skoltech}
\icmlauthor{Evgeny Burnaev}{Skoltech}
\end{icmlauthorlist}

\icmlaffiliation{Skoltech}{Skolkovo Institute of Science and Technology, Russia}
\icmlaffiliation{Criteo}{Criteo AI, France}
\icmlaffiliation{TII}{Technology Innovation Institute, UAE}

\icmlcorrespondingauthor{Gleb Bazhenov}{gleb.bazhenov@skoltech.ru}

\icmlkeywords{out-of-distribution detection, graph classification, uncertainty estimation, deep learning}

\vskip 0.3in
]



\printAffiliationsAndNotice{}  

\begin{abstract}

The problem of out-of-distribution detection for graph classification is far from being solved. The existing models tend to be overconfident about OOD examples or completely ignore the detection task. In this work, we consider this problem from the uncertainty estimation perspective and perform the comparison of several recently proposed methods. In our experiments, we find that there is no universal approach for OOD detection, and it is important to consider both graph representations and predictive categorical distribution.

\end{abstract}

\section{Introduction}

Out-of-distribution (OOD) detection requires us to distinguish between the in-distribution (ID) observations that can be processed by model and OOD samples that have to be rejected as inconsistent with the training data. This problem is out there for a long time for all kinds of modalities including the graph-structured data.




One way to deal with this problem is to provide an approach that assigns point uncertainties for a particular model.
In general, there are several approaches to uncertainty estimation in classification task that employ the ensembling techniques \cite{gal2016dropout, lakshminarayanan2017simple}, the probabilistic frameworks based on Dirichlet distribution \cite{malinin2018predictive, sensoy2018evidential, charpentier2020posterior}, and other methods \cite{liu2020simple, van2020simple, van2021improving, mukhoti2021deterministic}. While there were previous attempts to explore the OOD detection for image classification \cite{sinhamahapatra2022all}, their direct application to graph classification requires further research due to a complex network structure and additional attributes associated with the nodes.
\\

Recent works \cite{zhao2020uncertainty, stadler2021graph} consider the OOD detection in node classification problem using uncertainty estimation; however, the node-level detection is different from the graph-level one due to the dependent relationships between nodes, which does not occur for graphs. Likewise, some works \cite{wu2021handling, ding2021closer, li2022out} explored the question of OOD generalisation on graph-structured data, where robust models have been proposed to deal with the distribution shifts; however, such works do not consider the detection of OOD samples, which is the focus of our paper. 



The problem of OOD detection for graph classification is not solved yet. The existing models tend to be overconfident about OOD examples or completely ignore the detection task. In our work, we consider this problem from the uncertainty estimation perspective and perform the comparison of several recently proposed methods. In particular, we find that there is no universal approach for OOD detection across different graph datasets, and it is important to consider not only the entropy of predictive distribution, but also the geometry of the latent space produced by graph encoder. In addition, we observe that sometimes the classification task has a \textit{default} ID class for most OOD samples.





\section{Methods}
\label{methods}

In general, we construct the model that for a given graph $x$ and the associated classifier $c(x)$ can provide the uncertainty estimate $u(x)$. Since these estimates are used for ranking the test samples, we require that such model predicts higher uncertainty for OOD examples.

In this work, we consider two common distinct classes of methods for OOD detection through uncertainty estimation:
\begin{itemize}
    \item entropy-based methods, which use the entropy of predictive categorical distribution as the measure of uncertainty, treating the predictions with high entropy as less certain;
    
    
    \item density-based methods, which perform probability density estimation using the provided representations in latent space and assign high uncertainty to the points with low density.
\end{itemize}

The entropy-based methods expect that we have a particular classification model, while the density-based focus on a density estimate $\mathtt{p}(x)$ that can be independent from our classifier. However, in practice, they can share the representations with the main classification model.

Most methods that we describe here can distinguish between two sources of uncertainty \cite{gal2016uncertainty} — data uncertainty $u_{\text{data}}$, which reflects the internal noise in data due to class overlap or data labeling errors, and knowledge uncertainty $u_{\text{know}}$, which refers to the lack of knowledge about unseen data. In our experiments, we use both types of uncertainty and investigate which one is more consistent with the OOD detection problem.

\subsection{Entropy-Based}
\label{methods:entropy}


\begin{figure*}[t]
\centering
\caption{OOD detection performance per class of all the considered methods using \\ knowledge uncertainty KU $u_{\text{know}}$ on ENZYMES (left) and IMDB-MULTI (right) datasets.}
\includegraphics[keepaspectratio, height=5cm]{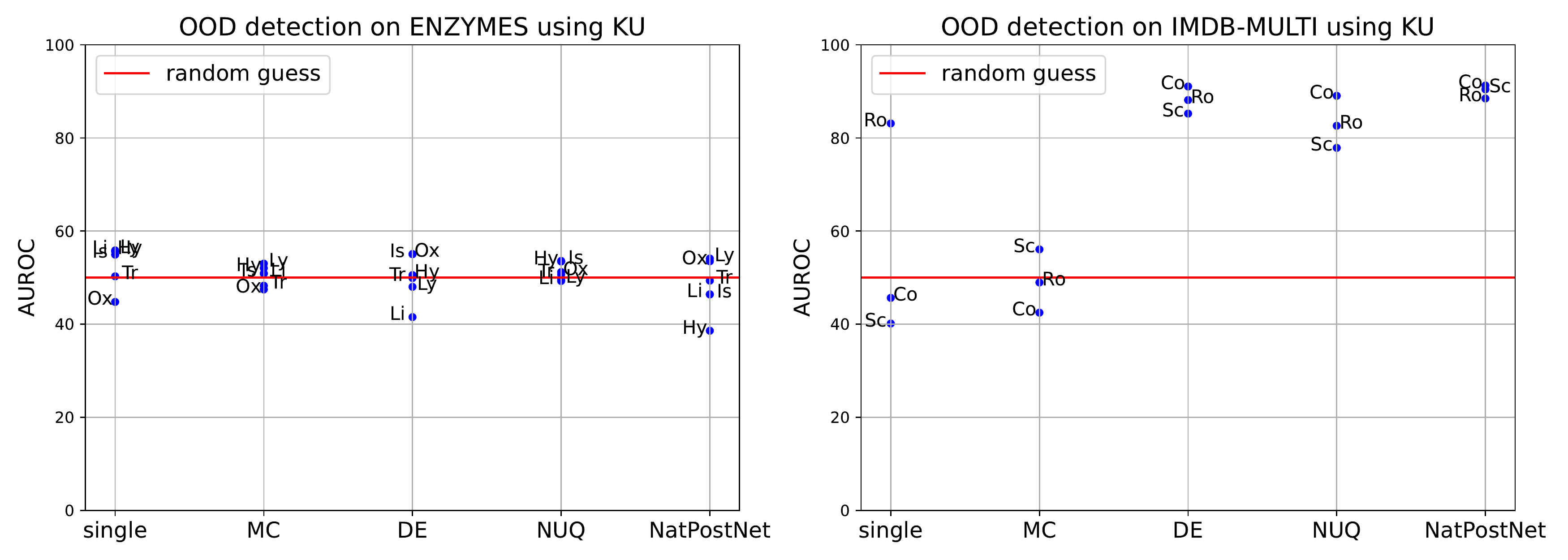}
\label{fig:ood-detection-main}
\end{figure*}

\begin{figure*}[h]
\centering
\caption{t-SNE graph embeddings produced by \texttt{single} for classes with 1 (left) and 5 (right) triangles \\ from TRIANGLES dataset. It reflects how the joint distribution of graph representations \\ changes as the OOD class moves from the boundary of ID space to its interior.}
\includegraphics[keepaspectratio, height=6.5cm]{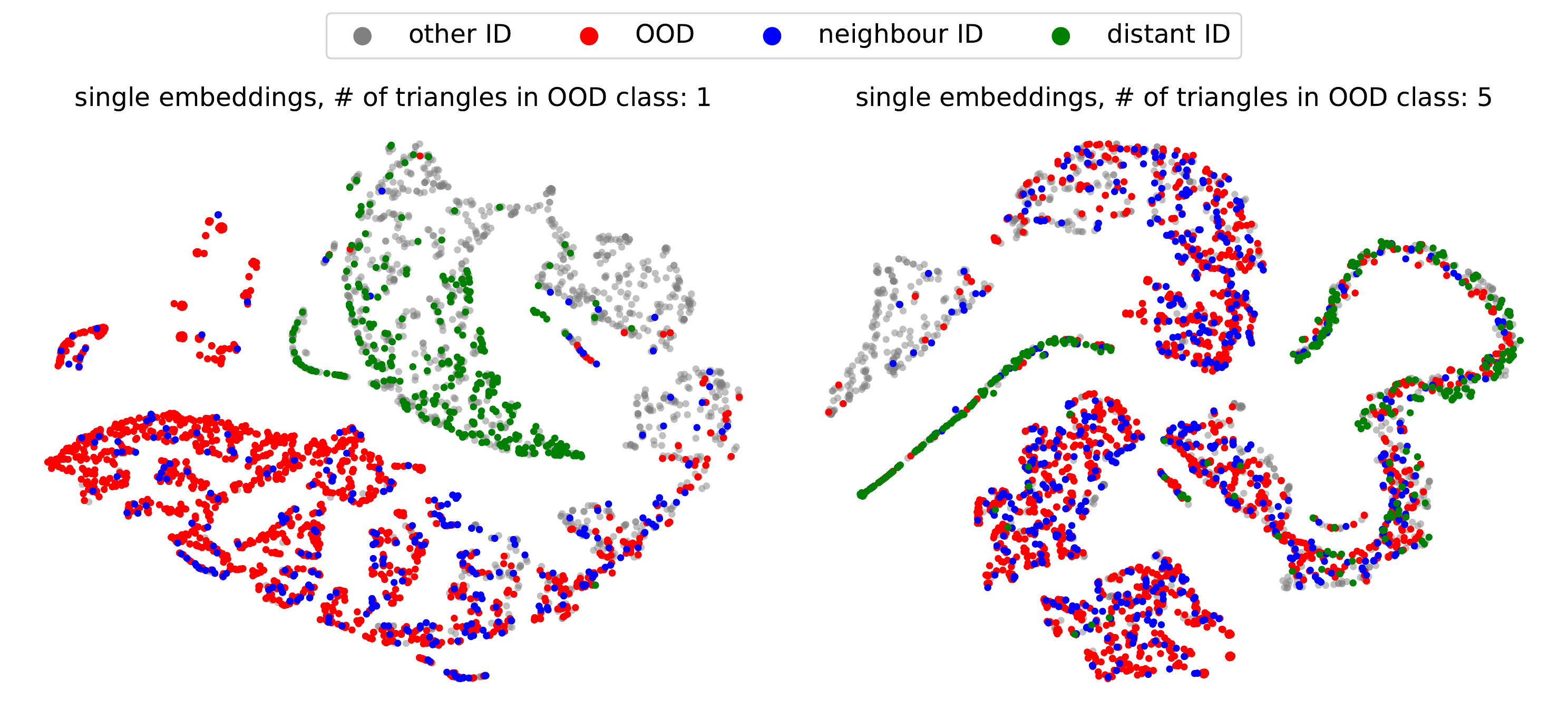}
\label{fig:tsne-main}
\end{figure*}

The entropy-based methods that we use in our experiments can be described in terms of the probabilistic framework which implies some unknown distribution $\mathtt{p}(\theta|\mathcal{D})$ of model parameters $\theta$ given the training data $\mathcal{D}$. In our classification task, the predictive distribution is expressed as follows:
\[
    \mathtt{P}(y|x, \mathcal{D}) = \mathbb{E}_{\mathtt{p}(\theta|\mathcal{D})}\mathtt{P}(y|x, \theta),
\]
where $\mathtt{P}(y|x, \theta)$ is defined by a graph classification model $f_{\theta}(x)$. Each method presented below provides a particular form of distribution $\mathtt{p}(\theta|\mathcal{D})$ and the uncertainty measure that is used for OOD detection in graph classification problem.

\subsubsection{Single GNN Model}
\label{methods:single}

This approach employs the most simple form of distribution $\mathtt{p}(\theta|\mathcal{D})$ which is expressed through a delta function around the MLE of model parameters $\theta$:
\[
    \mathtt{p}(\theta|\mathcal{D}) = \delta(\theta - \theta_{\text{MLE}}).
\]
In such case, the predictive distribution has the form 
\[
    \mathtt{P}(y|x, \mathcal{D}) = \mathtt{P}(y|x, \theta_{\text{MLE}}).
\]
The model consists of several graph convolution layers and global graph pooling that provides a representation $z(x)$, which is then processed by a single-layer fully-connected network in order to obtain the predictive distribution.

In this approach, we do not distinguish between data and knowledge uncertainty and base the total uncertainty measure of this model on the entropy of predictive distribution:
\[
    u_{\text{total}} = \mathbb{H}\big[\mathtt{P}(y|x, \theta_{\text{MLE}})\big].
\]
Further in the text, we refer to this method as \texttt{single}.

\subsubsection{Ensemble of GNN Models}
\label{methods:ensemble}


We also consider two ensembling techniques for constructing more robust predictive models — Monte Carlo Dropout \cite{gal2016dropout} and Deep Ensembles \cite{lakshminarayanan2017simple}. To formally define these approaches, we consider the approximation of distribution $\mathtt{p}(\theta|\mathcal{D})$ via a set of models, which either represent the dropout versions of the same estimator described in \cref{methods:single} (MC), or independently trained instances of such model (DE):
\[
    \theta \sim \mathtt{q}(\theta|\mathcal{D}),
\]
where the distribution $\mathtt{q}(\theta|\mathcal{D})$ is discrete uniform over $n$ model instances $\theta_k$. Given this, the predictive distribution can be expressed as a discrete average instead of the original integral:
\[
    \mathtt{P}(y|x, \mathcal{D}) = \frac{1}{n}\sum\limits_{k = 1}^n \mathtt{P}(y|x, \theta_k).
\]
It enables us to separate data and knowledge uncertainty in the same way as it was done in~\cite{malinin2018predictive}:
\[
\begin{split}
\begin{gathered}
    u_{\text{data}} = \mathbb{E}_{\mathtt{q}(\theta|\mathcal{D})} \mathbb{H}\big[\mathtt{P}(y|x, \theta)\big], \\
    u_{\text{know}} = \mathbb{H}\big[\mathbb{E}_{\mathtt{q}(\theta|\mathcal{D})} \mathtt{P}(y|x, \theta)\big] - u_{\text{data}}.
\end{gathered}
\end{split}
\]
%
For the sake of brevity, we refer to Deep Ensemble as \texttt{DE} and Monte Carlo Dropout as \texttt{MC}.

\subsection{Density-Based}
\label{methods:density}

These methods are rather more flexible and diverse in terms of formulation, but their main idea is to estimate the density function of probability distribution in the latent space of graph representations $z(x)$ which are associated with a particular classification task.
The density-based methods all suffer from challenging density estimation in high-dimensional embedding space. However, existing approaches in various ways try to mitigate this issue for large-dimension spaces.

\subsubsection{NUQ}
\label{methods:nuq}

The first density-based method for uncertainty estimation that we consider is Non-Parametric Uncertainty Quantifaction \cite{kotelevskii2022nuq}, which can be derived from the problem of statistical risk minimisation and leads to the natural decomposition of total uncertainty into the sum of data uncertainty
\[
    \eta_k(x) = \mathtt{P}(y = k|x) \Longrightarrow u_{\text{data}} = \min_k \eta_k(x),
\]
where $\mathtt{P}(y = k|x)$ is defined by the Nadaraya-Watson estimator \cite{nadaraya1964estimating}, and knowledge uncertainty
\[
    u_{\text{know}} \sim \max\limits_k \sigma_k^2(z) / \mathtt{p}(z),
\]
where $\sigma_k^2(z) = \eta_k(x)\big[1 - \eta_k(x)\big]$ and the probability density function $\mathtt{p}(z)$ is approximated using the kernel density estimator (KDE) in graph representation space, which in turn is induced by the already trained graph encoder described in \cref{methods:single}. We use its short notation \texttt{NUQ} for convenience.

\subsubsection{NatPostNet}
\label{methods:natpostnet}

Another view on the uncertainty estimation is given by Natural Posterior Network \cite{charpentier2021natural}, which predicts the parameters of Dirichlet distribution instead of directly modeling the posterior categorical distribution. 


Using the parameters of Dirichlet distribution $\mathtt{p}(\mu|x, \theta)$, the predictive categorical distribution can be defined as
\[
    \mathtt{P}(y|x, \mathcal{D}) = \mathbb{E}_{\mathtt{p}(\mu|x, \theta)}\mathtt{P}(y|\mu),
\]

where the distribution $\mathtt{p}(\mu|x, \theta)$ is learned using the same model $f_{\theta}(x)$ as described in \cref{methods:single} — the only difference is that such GNN predicts the Dirichlet parameters instead of categorical ones. In this framework, data uncertainty is defined as the entropy of predictive distribution:
\[
    u_{\text{data}} = \mathbb{H}\big[\mathtt{P}(y|x, \mathcal{D})\big].
\]

As for knowledge uncertainty, it can be expressed through the density estimate provided by a single normalising flow $g_{\psi}(z)$ \cite{kobyzev2020normalizing} over graph representations $z$:
\[
    u_{\text{know}} \sim -g_{\psi}(z).
\]

This method is denoted by \texttt{NatPostNet}.

\begin{figure*}[t]
\centering
\caption{OOD confusion (left) and pair-wise distance (right) matrices \\ produced by \texttt{NatPostNet} on REDDIT-MULTI-12K dataset.}

\begin{subfigure}
\centering
\includegraphics[keepaspectratio, height=7cm]{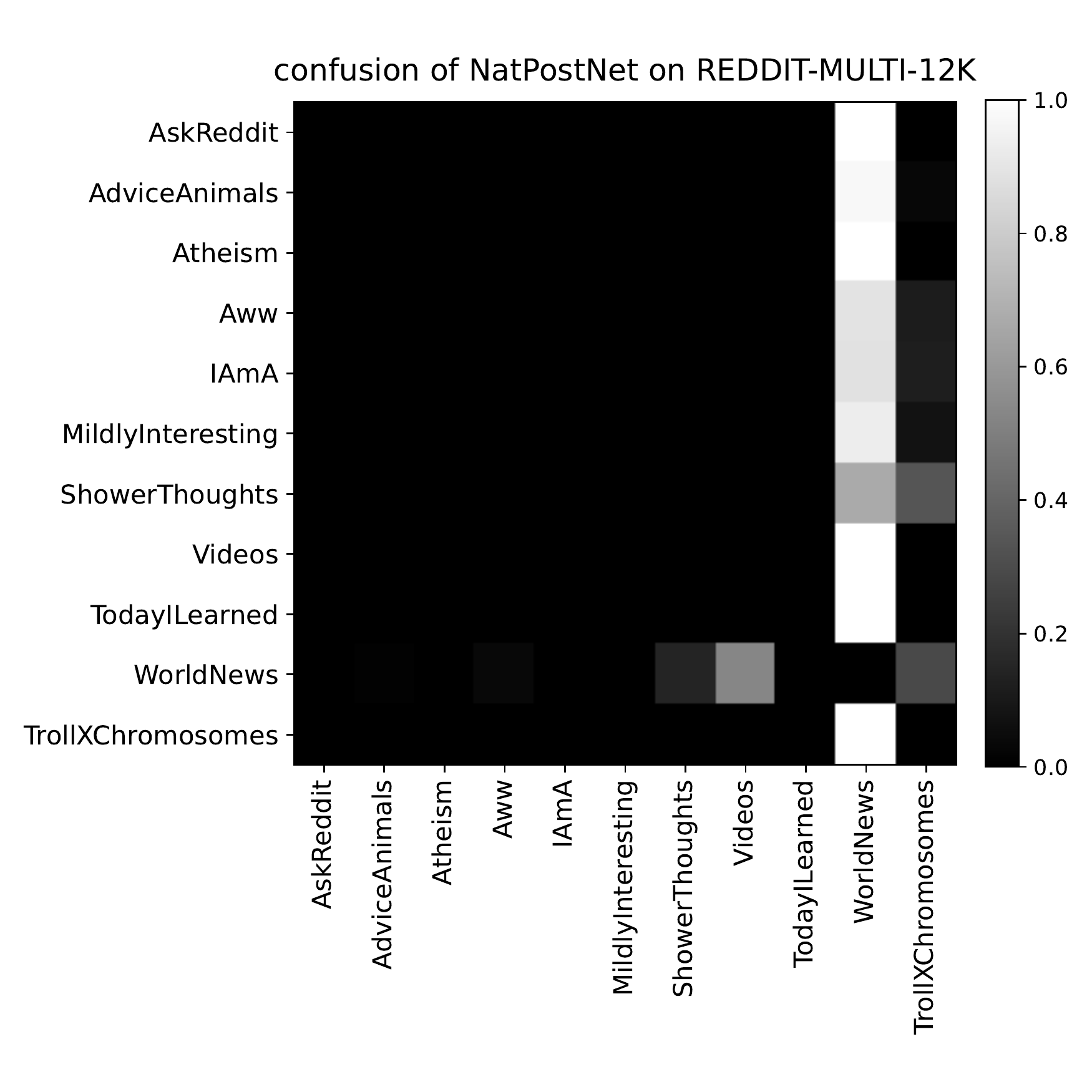}
\label{fig:confusion-main}
\end{subfigure}
\begin{subfigure}
\centering
\includegraphics[keepaspectratio, height=7cm]{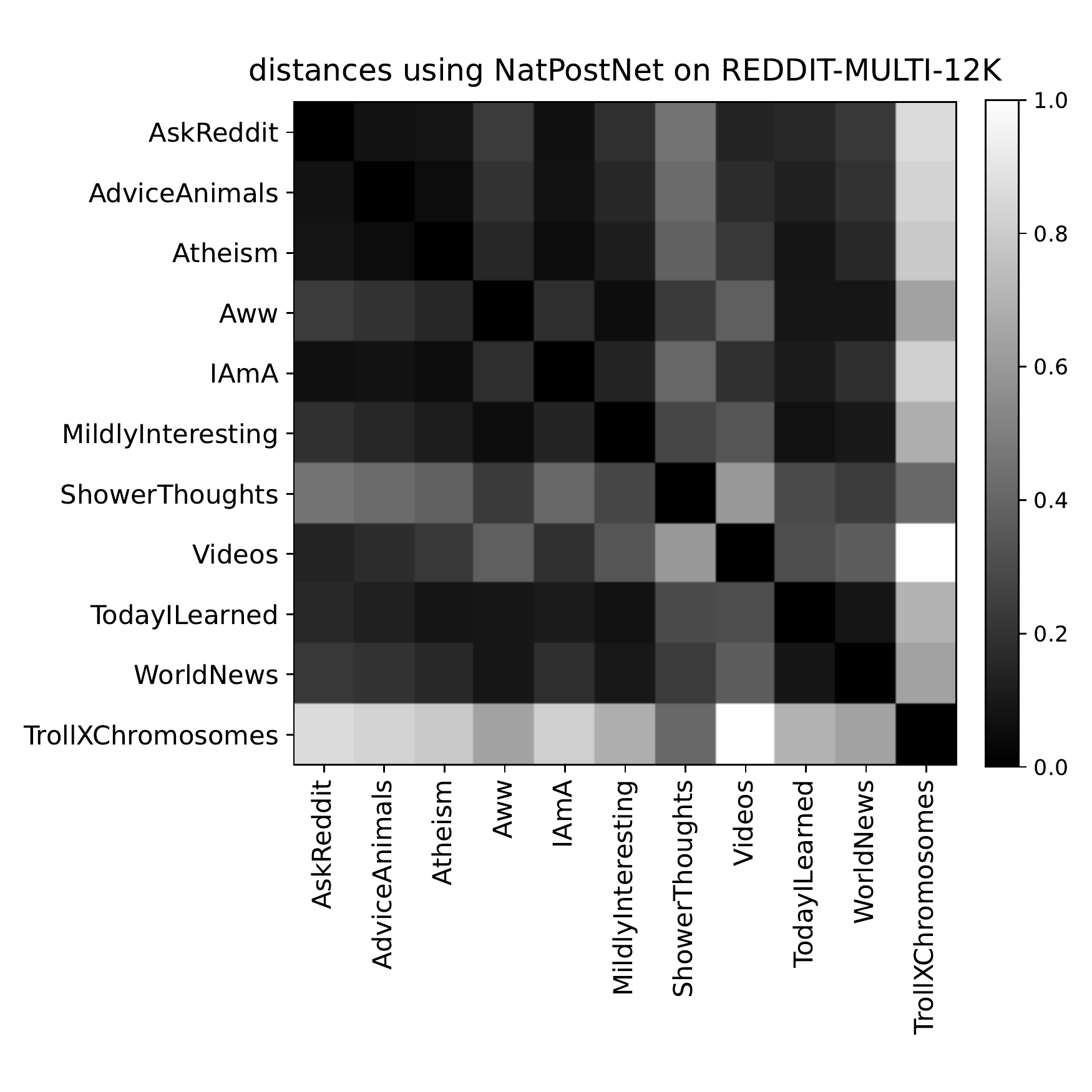}
\label{figure:distances-main}
\end{subfigure}

\label{fig:distances-confusion-main}
\end{figure*}

\section{Experiments}
\label{experiments}



Because of the space limitation, we report the graph dataset statistics and their class names in \cref{appendix:datasets}. Moreover, we describe the evaluation procedure and OOD detection metrics for all datasets in \cref{appendix:experiments,appendix:ood-performance}, while here we discuss only the key insights that we have managed to obtain during our experiments. In particular, we formulate and answer the following questions:

\textit{Are there methods that provide the optimal OOD detection performance on both chemical and social graph datasets?}

Based on \cref{fig:ood-detection-main}, it is natural to conclude that chemical and social datasets vary in the difficulty of OOD detection for the same graph encoder architecture, since the performance of \texttt{DE} and both considered density-based methods on IMDB-MULTI is much better than on ENZYMES. It means that particular applications such as chemistry might require a special design of encoders that model the graph manifold more efficiently.

    
\textit{How the OOD detection performance is influenced by the structure of the graph dataset and the associated latent space induced by the model?}

As it can be seen from the last row in \cref{fig:ood-detection-all}, the OOD detection performance of the entropy-based methods on TRIANGLES dataset drops significantly as the OOD class moves from the interior of ID region to its boundary (i.e., from 5 to 1 or 10). We can say that these methods become overconfident for the boundary classes.

However, looking on the distribution of graph embeddings (see \cref{fig:tsne-main}), one can notice that for the boundary class (1 triangle) the OOD embeddings seem to be better distinguishable from ID ones than for the classes in the middle of the range (5 triangles). It means that the usage of methods based solely on the entropy of predictive distribution is not enough, and one also may need to take into account the geometry of embeddings.


\textit{What the predictive categorical distribution over OOD samples can say about the ID classes?}

We recall that the classes that are available to classifier are referred to as ID and those that it has not seen are treated as OOD. By retraining the classifiers on a different set of ID classes we can obtain different predictions for OOD classes. In \cref{fig:confusion-main} (left), one can notice that regardless of a particular OOD class, the model often predicts it to be the \texttt{WorldNews} class, which can be treated as a \textit{default} ID class for any OOD sample.

    


\textit{Can we confirm that the most distant classes in the latent space appear to be the most simple for OOD detection?}

Comparing the OOD detection performance using knowledge uncertainty on REDDIT-MULTI-12K dataset in the corresponding row of \cref{fig:ood-detection-all} and pair-wise distances between its classes in \cref{fig:confusion-main} (right), one can understand that the most distant classes such as \texttt{TrollXChromosomes} tend to be easier for OOD detection.

\textit{What type of uncertainty is more consistent with the OOD detection problem for graph classification?}

In general, the estimates based on knowledge uncertainty appear to be more appropriate for OOD detection than using data uncertainty, as can be seen in \cref{fig:ood-detection-all}.


\section{Conclusion}

We investigated the OOD detection problem in graph classification task by comparing several uncertainty estimation methods and discussing some key insights about the dependence of OOD detection performance on the structure of graph representations and the interpretation of predictive categorical distribution in context of OOD detection.



\bibliography{references}
\bibliographystyle{icml2022}

\newpage
\appendix
\onecolumn

\section{Experimental Setup}
\label{appendix:experiments}


In this work, we consider a particular OOD detection setup when one of the originally available $C$ classes in graph dataset is removed and marked as OOD, while the graph classification model is trained on the remaining set of classes in stratified validation mode. After that, we predict the uncertainty measures for both ID and OOD samples separated previously. Based on this, we evaluate the ability of the considered methods to rank observations and detect the OOD samples — we compute the AUROC metric on the united set of predictions for ID validation samples and OOD samples, treating the OOD detection problem as formal binary classification task. In other words, we solve one distinct ID classification problem for each of $C$ OOD classes. For the sake of a consistent evaluation, we perform 5 random splits of the ID set into train and validation parts, and then average the obtained metrics.


Along with the OOD detection metrics, we also provide the related materials which might be helpful for interpretation. \cref{appendix:tsne} contains the figures of graph embeddings on TRIANGLES dataset that represent the structure of latent space in the corresponfing OOD detection problem. Moreover, \cref{appendix:confusion} presents the OOD confusion matrices for all the considered datasets, where rows correspond to particular OOD classes, while columns — to the predicted ID classes. In \cref{appendix:distances}, we place the pair-wise distance matrices between classes in standard ID classification, which reveals the correlation between the distance to other classes and the difficulty of OOD detection.

\section{Datasets Details}
\label{appendix:datasets}

\subsection{Basic Statistics}
\label{appendix:statistics}

In \cref{tab:datasets}, we report basic graph dataset statistics, such as number of graphs and avergae number of nodes and edges.

\begin{table}[!ht]
\centering
\caption{Dataset statistics. Note that only ENZYMES dataset has natural features \\ in nodes, so we assign the same artifiial feature 1 to nodes in other datasets.}
\label{tab:datasets}
\begin{tabular}{lccccc}
\toprule
Name & \# Graphs & \# Classes & \# Nodes & \# Edges & \# Features \\
\midrule
ENZYMES          &         600 &            6 &      32.63 &      62.14 &             3 \\[1pt]
IMDB-MULTI       &        1500 &            3 &      13.00 &      65.94 &             1 \\[1pt]
REDDIT-MULTI-5K  &        4999 &            5 &     508.52 &     594.87 &             1 \\[1pt]
REDDIT-MULTI-12K &       11929 &           11 &     391.41 &     456.89 &             1 \\[1pt]
TRIANGLES        &       45000 &           10 &      20.85 &      32.74 &             1 \\
\bottomrule
\end{tabular}
\end{table}

\subsection{Class Names}
\label{appendix:classes}

In \cref{tab:class-names}, you can find the list of class names for each considered graph dataset.

\begin{table}[!ht]
\centering
\caption{List of class names for every considered graph dataset.}
\label{tab:class-names}
\begin{tabular}{ll}
\toprule
Name & Class Names \\
\midrule
ENZYMES          & \texttt{Oxidoreductases Transferases Hydrolases} \\
                 & \texttt{Lyases Isomerases Ligases}
\\[8pt]
IMDB-MULTI       & \texttt{Comedy Romance Sci-Fi}                                                
\\[8pt]
REDDIT-MULTI-5K  & \texttt{WorldNews Videos AdviceAnimals} \\
                 & \texttt{Aww MildlyInteresting}
\\[8pt]
REDDIT-MULTI-12K & \texttt{AskReddit AdviceAnimals Atheism Aww IAmA} \\
                 & \texttt{MildlyInteresting ShowerThoughts Videos} \\
                 & \texttt{TodayILearned WorldNews TrollXChromosomes}
\\[8pt]
TRIANGLES        & \texttt{1 2 3 4 5 6 7 8 9 10} \\
\bottomrule
\end{tabular}
\end{table}

\newpage

\section{Out-of-Distribution Detection Performance}
\label{appendix:ood-performance}

The OOD detection metrics presented here are partially discussed in \cref{experiments} — they enable us to compare the performance of various methods and conclude that there is no universal approach for OOD detection in graph classification.

\begin{figure}[!h]
\centering
\caption{OOD detection performance per class of all the mentioned methods using \\ data uncertainty DU (left column) and knowledge uncertainty KU (right column) on all the considered datasets.}
\includegraphics[keepaspectratio, height=19.5cm]{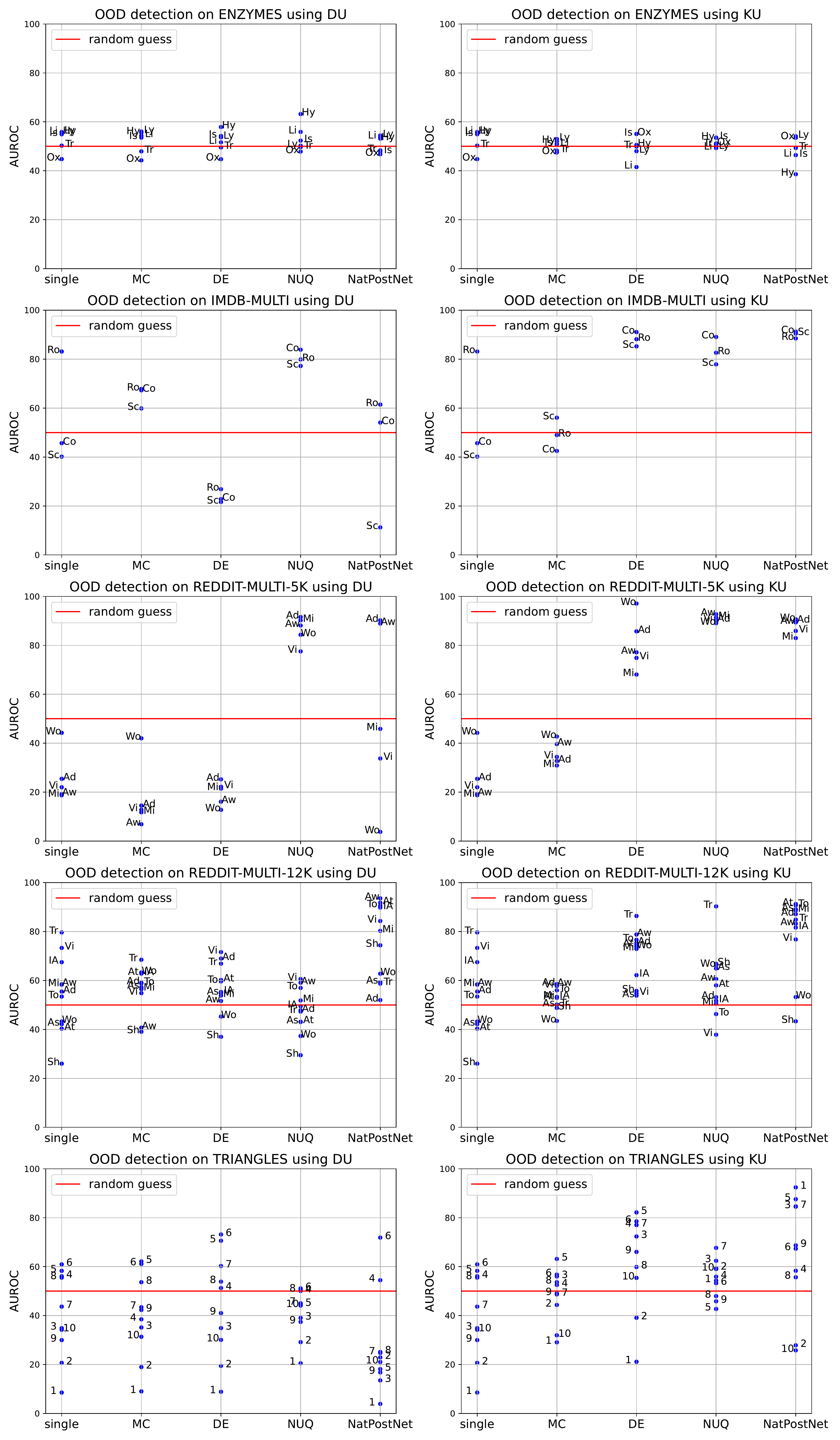}
\label{fig:ood-detection-all}
\end{figure}

\section{t-SNE Graph Embeddings}
\label{appendix:tsne}

Here you can see how the joint distribution of graph representations changes depending on the particular OOD class.

\begin{figure}[!h]
\centering
\caption{t-SNE graph embeddings produced by \texttt{single} for all classes from the TRIANGLES dataset. \\ Red points represent OOD graphs, blue points — graphs from the nearest ID class, \\ green — graphs from the most distant ID class, and grey — all other graphs.}
\includegraphics[keepaspectratio, height=19.5cm]{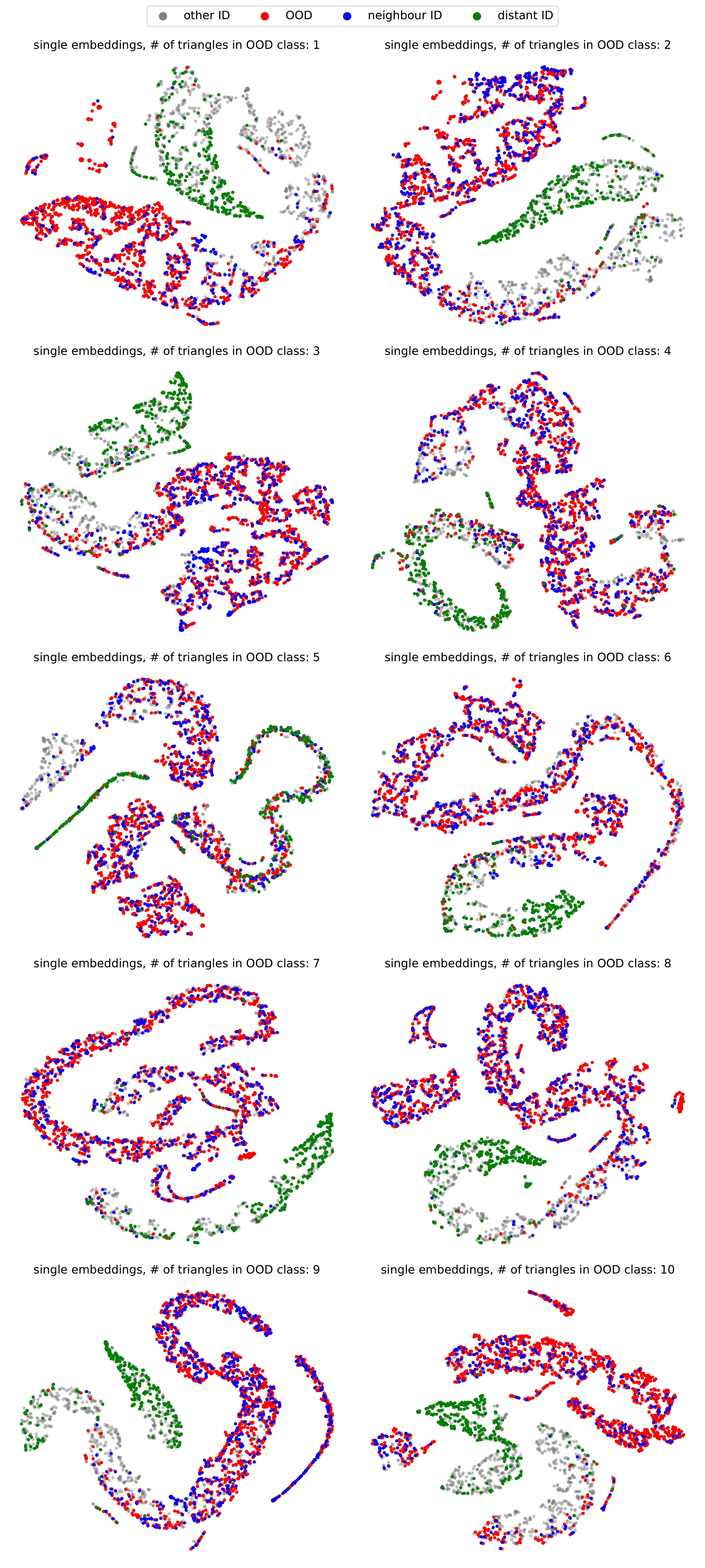}
\label{fig:tsne-all}
\end{figure}

\section{OOD Confusion Matrices}
\label{appendix:confusion}

These OOD confusion matrices are used to interpret the predictive categorical distribution over OOD samples and find that sometimes the classification task has a \textit{default} ID class for most OOD samples, as discussed in \cref{experiments}.

\begin{figure}[!h]
\centering
\caption{OOD confusion matrices produced by \texttt{single}, \texttt{DE} and \texttt{NatPostNet} for all the considered datasets. \\ Rows correspond to the OOD classes, while columns — to the predicted ID classes.}
\includegraphics[keepaspectratio, height=19.5cm]{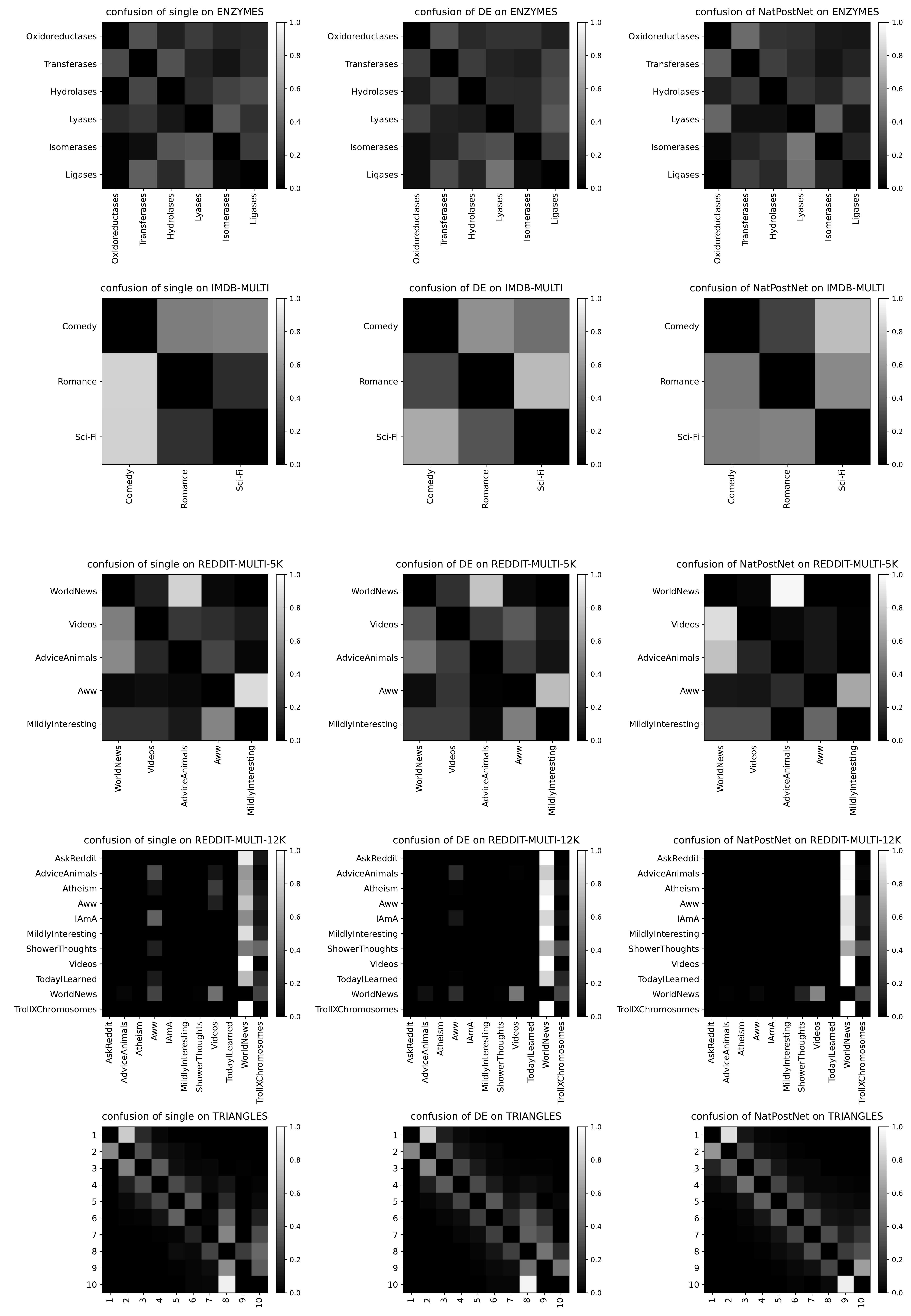}
\label{fig:confusion-all}
\end{figure}

\section{Distance Matrices Between Classes}
\label{appendix:distances}

Comparing these matrices and the OOD detection performance on the corresponding dataset, we can confirm that the most distant classes in latent space appear to be the most simple for OOD detection.

\begin{figure}[!h]
\centering
\caption{Distance matrices produced by \texttt{single} and \texttt{NatPostNet} for all the considered datasets. \\ The distance between classes is defined as the Euclidean distance between the corresponding centroids of graph embeddings.}
\includegraphics[keepaspectratio, height=19.7cm]{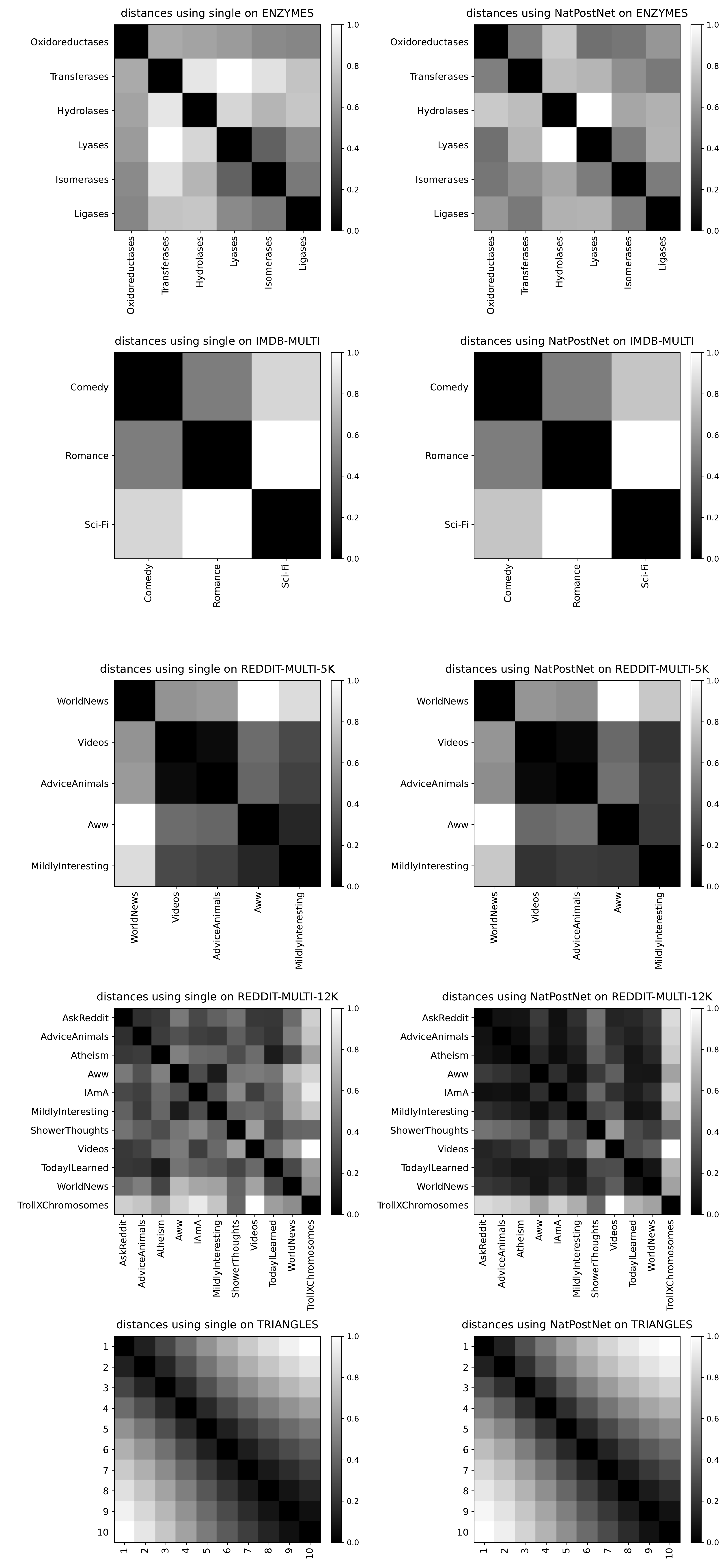}
\label{fig:distances-all}
\end{figure}


\end{document}